\documentclass{article} % For LaTeX2e
\usepackage{iclr2026_conference,times}
\iclrfinalcopy
\usepackage{amsmath,amsfonts,bm}

\def\eqref#1{equation~\ref{#1}}
\def\1{\bm{1}}

\DeclareMathAlphabet{\mathsfit}{\encodingdefault}{\sfdefault}{m}{sl}
\SetMathAlphabet{\mathsfit}{bold}{\encodingdefault}{\sfdefault}{bx}{n}

\def\gG{{\mathcal{G}}}

\usepackage{hyperref}
\usepackage{url}
\usepackage{graphicx}
\usepackage{multirow}
\usepackage{algorithm}
\usepackage{algpseudocode}

\newcommand{\OURS}{ProcGen3D}

\title{\OURS: Learning Neural Procedural Graphs for Image-to-3D Reconstruction}

\author{Xinyi Zhang$^1$, Daoyi Gao$^1$, Naiqi Li$^2$, Angela Dai$^1$
\\
$^1$ Technical University of Munich\\
$^2$ Tsinghua University \\
\texttt{xinyi.zhang@tum.de,\quad daoyi.gao@tum.de,\quad linaiqi.thu@gmail.com,}\\
\texttt{angela.dai@tum.de }  
}

\begin{document}

\maketitle

%%%%%%%%% ABSTRACT
\begin{abstract}
We introduce \OURS, a new approach for 3D content creation by generating procedural graph abstractions of 3D objects, which can then be decoded into rich, complex 3D assets. 
%Procedural generators are widely used in computer graphics for movies, video games, and engineering design of real-world structures. However, specifying such models and tuning their parameters remains challenging. 
Inspired by the prevalent use of procedural generators in production 3D applications, we propose  a sequentialized, graph-based procedural graph representation for 3D assets. 
We use this to learn to approximate the landscape of a procedural generator for image-based 3D reconstruction.
%is both geometrically expressive and compact, enabling the modeling of complex objects. To infer such graphs from observations, we propose a transformer-based neural inverse procedural modeling approach that predicts procedural graphs directly from RGB inputs. 
We employ edge-based tokenization to encode the procedural graphs, and train a transformer prior to predict the next token conditioned on an input RGB image. 
Crucially, to enable better alignment of our generated outputs to an input image, we incorporate Monte Carlo Tree Search (MCTS) guided sampling into our generation process, steering output procedural graphs towards more image-faithful reconstructions. 
Furthermore, this enables improved generalization on real-world input images, despite training only on synthetic data.
%To improve consistency between predictions and observations, we integrate Monte Carlo Tree Search (MCTS) into the generation process, guiding the model toward geometry-aware reconstructions.
Our approach is applicable across a variety of objects that can be synthesized with procedural generators. 
Extensive experiments on cacti, trees, and bridges show that our neural procedural graph generation outperforms both state-of-the-art generative 3D methods and domain-specific modeling techniques.
%We evaluate our approach across diverse procedural generators and demonstrate strong generalization to real-world objects.
\end{abstract}

% %%%%%%%%% BODY TEXT
\section{Introduction}

Creating immersive 3D content is in significant demand, with widespread applications in industries such as film, gaming, mixed reality, architecture, robotics, and more. 
This has spurred on significant advances in generative 3D object modeling \citep{xiang2025structured,chen20253dtopia}, which have shown remarkable promise. 
However, these methods typically generate outputs as neural fields \citep{zhang20233dshape2vecset,xiang2025structured}, points \citep{xie2021generative}, or voxels \citep{ren2024xcube,meng2025lt3sd} that must then be converted into meshes for downstream applications, typically through Marching Cubes \citep{lorensen1998marching} -- which results in over-tessellated meshes often lacking sharp geometry.
In contrast, procedural modeling has 
%Procedural generators have 
long been a cornerstone in computer graphics, powering the creation of rich and diverse 3D content across many production applications \citep{muller2006procedural, parish2001procedural, raistrick2023infinite}. 
By encoding domain-specific rules and stochastic processes, they can synthesize highly complex structures (e.g., plants, terrain, or architectural designs) with remarkable realism while maintaining compact encodings. 
%Beyond entertainment, procedural modeling also plays a key role in engineering and architecture, where compact generative rules enable the scalable design of intricate real-world structures.
This offers a compelling alternative in generative 3D modeling, to synthesize lightweight, parametric representations that lead to complex, sophisticated geometric models as output.

Despite their expressive power, obtaining a desired procedural model based on specified user intent (e.g., image-based 3D reconstruction) remains a significant challenge -- as the representation space is comparatively small while mixing continuous parameters with discrete choices. 
%procedural generators are difficult to specify and control. Designing rules and tuning parameters typically require expert knowledge and significant manual effort. 
Additionally, procedural models typically  involve stochastic sampling, which makes the relationship between input parameters and the final output nondifferentiable. 
This makes inverse procedural modeling through direct optimization very difficult.
%not only hinders optimization but also complicates inverse procedural modeling, i.e., recovering the underlying generative program from observed data.

\begin{figure}[t]
\begin{center}
\includegraphics[width=0.98\textwidth]{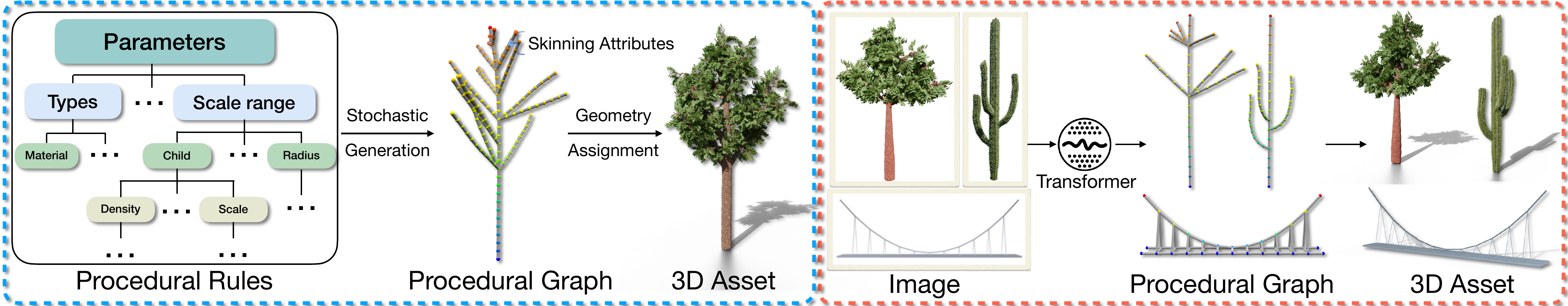}
\end{center}
\vspace{-0.4cm}
\caption{Left: procedural generators employ rule-based generation with stochastic sampling to produce abstract graph representations that are decoded into high-fidelity 3D assets through geometry and material assignment. 
Right: We propose to leverage such procedural graph representations, modeling their distribution with a transformer to enable high-fidelity image-to-3D reconstruction for various categories of procedurally generated objects (cacti, trees, bridges).}
\vspace{-0.4cm}
\label{fig:graph}
\end{figure}

To address these challenges, we propose to approximate the landscape of a procedural generator through a transformer-based neural network to encode a compact, abstract procedural graph representation for various 3D content, as shown in Fig.~\ref{fig:graph}. 
%To address these challenges, we introduce a procedural graph representation that compactly encodes 3D assets generated by procedural rules. 
Instead of storing only surface geometry, we abstract a 3D object into a graph whose nodes and edges capture its structural skeleton, together with spatial and geometric attributes. Such graphs are generated according to the underlying  rules of a procedural generator, providing a concise and interpretable description of the object. % which can then be decoded into complex geometric meshes
The flexible and expressive graph representation allow our method capture the skeleton and attributes of 3D content in a way that is interpretable and structurally valid.

% \begin{figure}[t]
% \begin{center}
% \includegraphics[width=0.95\textwidth]{figs/graph1.png}
% \end{center}
% \caption{General pipeline of procedural generators. }
% \label{fig:graph}
% \end{figure}

% Compared to conventional 3D representations—such as meshes, voxels, or point clouds—procedural graphs offer both compactness and structural validity, while enabling efficient control over variations.
In contrast to 3D voxel, point, or neural field representations, using such a procedural generation basis enables capturing complex 3D content in a very compact form that can be decoded into rich 3D assets with complex, sharp detail that are directly suitable for downstream production applications. 
%Traditional 3D representations—such as meshes, voxels, and point clouds—primarily capture the geometric shape of an object while ignoring the underlying generative or physical rules. This limitation often leads to structurally inconsistent or physically implausible objects, particularly for highly complex categories such as plants or categories that need to be physically grounded (e.g. architecture). In contrast, procedural graph representations explicitly encode both structure and geometry, enabling more faithful and interpretable modeling.

% As a result, structural correctness is not guaranteed and physically implausible structures are common in their generations.

Modeling such procedural graphs requires handling both their topological structure as well as dependencies between edges and attributes. Inspired by the success of autoregressive transformers~\citep{vaswani2017attention} in language and multimodal modeling, we adopt a GPT-style transformer to learn procedural graph generation. Specifically, we sequentialize each graph into edge-based tokens, where each token encodes the spatial positions of its two endpoint vertices, their associated attributes, and the attributes of the edge.
% where each token encodes the attributes of its two endpoint vertices together with those of the edge.
% where each token encodes the attributes of its two nodes and the next connecting edge \ANGIE{what does connecting edge mean here exactly?}. 
The transformer is then trained to autoregressively for next-token prediction. %predict the next token conditioned on both the previously generated tokens and visual features extracted from the corresponding RGB image.

We adopt this approach for image-based 3D reconstruction, conditioning our graph-based transformer on input image features. 
However, while this can produce output procedural graphs that can capture the overall structure of an object, this often struggles to reflect accurate fine-grained structural details of the input image (e.g., the precise growth direction of a cactus stem following the input image). 
% \ANGIE{could improve this example}. 
This is partly due to the stochasticity of procedural generation and the vast combinatorial space of possible graphs. As a result, relying solely on autoregressive prediction from image features can lead to inconsistencies or loss of detail in the reconstructed graphs.

We thus propose to further integrate Monte Carlo Tree Search (MCTS) \citep{metropolis1949monte,browne2012survey} guided sampling into our inference process. 
We leverage MCTS to explore multiple candidate continuations of the token sequence, based on our learned transformer prior, aiming to balance exploration of new possibilities with exploitation of high-scoring predictions. By evaluating simulated sequences against the image condition and propagating the feedback through the search tree, MCTS refines the transformer’s outputs and guides the generation toward procedural graphs that are both structurally consistent and more geometrically faithful to an input image.
Additionally, our test-time search enables our approach, trained only on synthetic data where ground truth procedural graph information is available, to generalize to real images inputs.
Our approach is versatile across various types of objects that can be modeled with procedural generators, which we demonstrate with experiments on three diverse categories: cacti, trees, and bridges. 

The main contributions of our work are summarized as follows:
\begin{itemize}
\item{
We propose to learn a neural procedural graph-based representation for 3D assets, which enables modeling abstract procedural graphs that can decode to rich, complex 3D assets. % that is rich in geometry and could represent complex objects compactly.
} 

\item {
Our approach enables neural inverse procedural modeling, conditioned on RGB images as input to predict procedural graph abstractions for various object categories. 
}

\item {
To improve adherence to the input images and better generalize to real image inputs, we introduce image guidance with Monte Carlo tree search (MCTS) during our generation process, enabling sampling from our learned prior  while improving image consistency.
}
\end{itemize}

\section{Related Work}

\vspace{-0.2cm}
\paragraph{\textbf{Learning-Based 3D Object Generation.}}
Learning-based 3D generative methods have seen significant advances in recent years. 
Early approaches leveraged generative adversarial learning to train neural networks for unconditional 3D object synthesis~\citep{3dgan,gao2022get3d}.
Following the success of denoising diffusion models for 2D image generation, diffusion modeling has been widely adopted for 3D generation of various representations, including voxels \citep{hui2022neural,tang2023volumediffusion,ren2024xcube,meng2025lt3sd}, point clouds \citep{zhou20213d,luo2021diffusion}, and neural field representations \citep{erkocc2023hyperdiffusion,shue20233d,xiang2025structured}.
These methods have shown remarkable potential in synthesizing visually compelling 3D objects; however, they rely on Marching Cubes \citep{lorensen1998marching} post-processing to convert outputs into meshes for downstream applications, resulting in over-tessellated and over-smoothed geometric outputs.
Similarly, the success of autoregressive transformers in natural language modeling has also inspired the use of GPT-style transformers for mesh generation \citep{siddiqui2024meshgpt,chen2024meshanything}.
These methods can produce much more compact, sharper mesh outputs, but are limited in sequence context to low face count outputs.
Our approach also leverages GPT-style autoregressive transformers, but rather than operate on lengthy mesh sequences, we focus on modeling procedural graph representations.
This representation is inherently more compact and expressive, allowing later decoding into sophisticated, high-fidelity 3D assets.

\vspace{-0.2cm}
\paragraph{\textbf{Image-to-3D Generation.}}
For many downstream tasks, user input to control the generation process plays a key role. 
In particular, input RGB images provide a simple and very accessible way to guide 3D object generation.
Various methods have been proposed to leverage diffusion to generate neural fields from input images, representing output 3D shape geometry \citep{zhang20233dshape2vecset} and 3D radiance fields \citep{muller2023diffrf,liu2023zero,qian2023magic123,long2024wonder3d,xiang2025structured}, along with accelerated generation through transformer training \citep{hong2023lrm}.
Tree-D Fusion~\citep{lee2024tree} also introduces a domain-specific model leveraging diffusion priors, in order to synthesize trees from image inputs.
A more prevalent approach has been to leverage large shape databases for training; for instance, TRELLIS~\citep{xiang2025structured} employs latent flow matching to show remarkable 3D shape synthesis results from images. 
However, these methods all rely on Marching Cubes-style post-processing to obtain mesh outputs, which can introduce noise and oversmoothing in the output geometry. 
Furthermore, while such diffusion-based models can generated 3D shapes from an image, their outputs are primarily guided by a learned prior, and may not remain consistent with the input image, especially in the local structure.
In contrast, we employ not only a compact, procedural graph representation, but moreover, we propose a MCTS-guided test-time search to synthesize outputs that align well with the image input. 

\vspace{-0.2cm}
\paragraph{\textbf{Inverse Procedural Modeling.}}
Though procedural modeling \citep{muller2006procedural, parish2001procedural, raistrick2023infinite} has been widely used to generate high-quality 3D assets, the inverse problem -- recovering procedural representations from observed data --  remains highly challenging due to the combinatorial search space and the nondifferentiable nature of many procedural generators.
Early work on inverse procedural modeling primarily focused on recovering procedural rules through optimization and grammar induction of one specific category of objects, such as trees \citep{vst2010inverse} or buildings \citep{wu2013inverse}. These methods typically assume a predefined grammar, such as L-systems or split grammars, and attempt to infer rule parameters or production rules that best explained the observed data. Approaches in this category often relied on handcrafted heuristics, probabilistic inference, or search-based strategies to align generated procedural structures with target observations \citep{martinovic2013bayesian}. While effective for relatively constrained single domains such as plants or architectural facades, they tend to be sensitive and struggle to scale to highly complex or noisy real-world data.

Recently, several learning-based approaches have been proposed to overcome limitations of purely grammar-driven methods. These methods leverage neural networks to regress procedural parameters directly from input observations or to approximate procedural generators with differentiable surrogates, enabling gradient-based optimization \citep{plocharski2024faccaid}.
Generation-based approaches have also been explored, directly generating procedural  parameters from images using conditional diffusion models \citep{zhao2025di}. While effective, these methods are primarily suited for objects with relatively simple structures. In contrast, our method targets more complex categories by recovering full procedural graphs that explicitly capture both structural skeleton and geometric attributes.
We further demonstrate its versatility across diverse output types, including cacti, trees, and bridges.

\section{Method}

\begin{figure}[t]
\begin{center}
\includegraphics[width=0.90\textwidth]{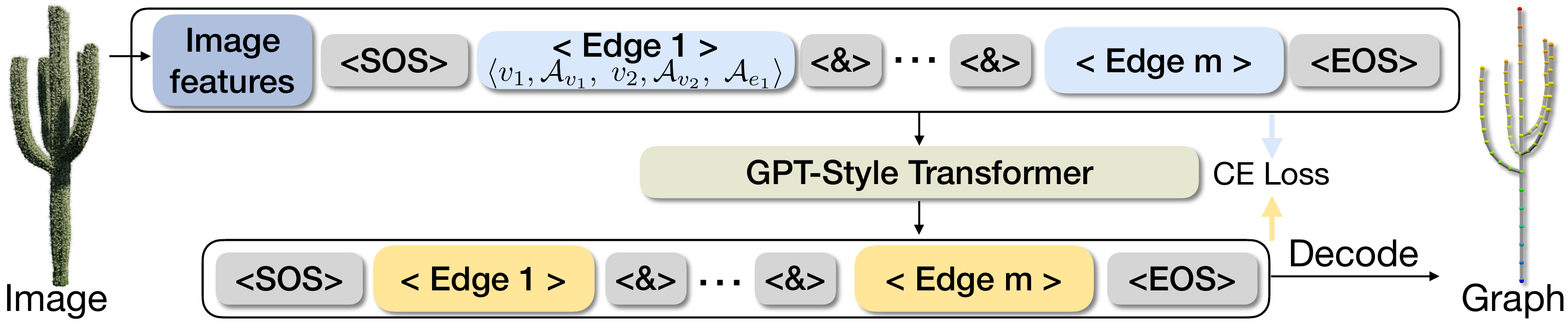}
\end{center}
\vspace{-0.4cm}
\caption{Overview of our graph-based transformer. A procedural graph is tokenized into a sequence of edge-based tokens, where each token encodes the positions and attributes of its two endpoint vertices as well as the attributes of the edge itself. The transformer autoregressively predicts tokens conditioned on image features, enabling reconstruction of the underlying procedural graph.}
\end{figure}

% \ANGIE{can you check that i introduced the main variables here correctly}
We introduce our procedural graph-based 3D representation $\displaystyle \gG$, which encodes a 3D object as a set of nodes and edges, enriched with geometric and semantic attributes. We then use this representation for image to 3D reconstruction by training a transformer model $\mathcal{T}$ to predict the underlying procedural graph from an RGB image observation. To facilitate faithful reconstructions to the input image, we further introduce MCTS-guided sampling of $\mathcal{T}$ at test time. 

% \begin{figure}[t]
% \begin{center}
% \includegraphics[width=0.95\textwidth]{figs/procedural_graph.png}
% \end{center}
% \caption{General pipeline of procedural generators. }
% \end{figure}

\subsection{Procedural Graph Representation}
Inverse procedural modeling is often formulated as recovering the initial parameters of a procedural generator; however, the non-differentiable stochastic sampling during generation makes this very difficult to optimize directly.
Instead, we propose to approximate the landscape of procedural generation by training a transformer on an intermediate procedural graph representation $\displaystyle \gG$.
%Previous work on inverse procedural modeling has primarily focused on recovering the initial parameter inputs of the procedural generator. However, the generation process often involves stochastic sampling, which makes it non-differentiable and difficult to optimize directly. 
%In fact, between the initial parameter set and the final 3D output, we can extract an intermediate graph representation that encodes both the structural skeleton and the associated spatial and geometric attributes on its nodes and edges.

The graph $\displaystyle \gG$ encodes the procedural information, capturing both structural skeleton information along with spatial and geometric attributes in its nodes and edges. This enables modeling a tractable and expressive basis for modeling and reconstructing complex 3D assets from image observations. 
$\displaystyle \gG$ is then represented as a graph with  $n$ nodes and $m$ edges, where each node and edge is associated with a set of attributes: 
%With a procedural generator, a 3D object could be represented as a graph $\displaystyle \gG$ with $n$ nodes and $m$ edges, where each node and edge is associated with a set of attributes:  
\begin{equation}
    % \displaystyle \gG = (V, E, \mathcal{A}_V, \mathcal{A}_E, \mathcal{L}_V, \mathcal{L}_E),
    \displaystyle \gG = (V, E, \mathcal{A}_V, \mathcal{A}_E)
\end{equation}
where $V = \{v_1, \dots, v_n\}$ denotes the set of nodes, $E = \{e_1, \dots, e_m\} \subseteq V \times V$ is the set of edges, $\mathcal{A}_V$ the node attributes (e.g., class label, radius), and $\mathcal{A}_E$ the edge attributes (e.g., length, force, semantic type, etc.).
% $\mathcal{L}_V: V \rightarrow \mathcal{A}_V, \mathcal{L}_E: E \rightarrow \mathcal{A}_E$, label functions.

From a graph $\displaystyle \gG$, the procedural generator can then assign detailed geometry along with material and texture to the mesh surface, producing watertight, manifold meshes with clean topology, which is difficult to achieve with voxel or neural field based 3D representations. 
With this compact representation, a complex tree mesh with more than 10 million faces can be abstracted as a graph with only a few hundred nodes, greatly reducing both computation and storage costs.

%Moreover, the procedural generator naturally supports controlled variation, allowing one to easily generate variations with specific constraints, such as trees with varying levels of foliage or different leaf types, thereby offering explicit control over the generated objects.
%Angie: we don't really demonstrate anything to do with control so would not emphasize it here

\subsection{Modeling Procedural Graphs with a Transformer}
Inspired by the success of multimodal large language models, we use an autoregressive transformer to model procedural graphs. 
We first convert each graph into a sequence of edge-based tokens.

% Given a graph $\gG$ with $m$ edges, we treat each edge as a unit for tokenization. 
% For an edge connecting vertices $v_a$ and $v_b$, we represent its token as
% \begin{equation}
%     \tau(e_i) = \big(v_a, \mathcal{L}_v({v_a}), v_b, \mathcal{L}_v({v_b}), \; \mathcal{L}_e({e_i}) \big),
% \end{equation}
% where $\mathcal{A}_{v_1}$ and $\mathcal{A}_{v_2}$ denote the attribute sets of the two endpoint vertices, 
% and $\mathcal{A}_e$ represents the attributes associated with the edge itself.
% Note that for simplicity, as we will describe our method in terms of tokenized edges, we will use $e_i$ to denote $\tau(e_i)$.

Given a graph $\gG$ with $m$ edges, we treat each edge as a unit for tokenization. 
For an edge connecting vertices $v_a$ and $v_b$, we represent its token as
\begin{equation}
    \tau(e_i) = \big(v_a, \mathcal{A}_{v_a}, \; v_b, \mathcal{A}_{v_b}, \; \mathcal{A}_{e_i} \big),
\end{equation}
where $\mathcal{A}_{v_a}$ and $\mathcal{A}_{v_b}$ denote the attribute sets of the two endpoint vertices, 
and $\mathcal{A}_{e_i}$ represents the attributes associated with the edge itself.
Note that for simplicity, as we will describe our method in terms of tokenized edges, we will use $e_i$ to denote $\tau(e_i)$.

For the entire graph, we tokenize all edges in a predefined spatial traversal order. 
Specifically, we use a depth-first search (DFS) order for plant-like structures 
and a breadth-first search (BFS) order for architectures such as bridges. 
Between consecutive edge tokens, we insert a special split token $\langle \& \rangle$ to explicitly indicate edge boundaries, producing our final tokenized sequence as:
%\begin{equation}
 $   \mathcal{S}(\gG) = \{ e_1, \; \langle \& \rangle, \; e_2, \; \langle \& \rangle, \; \dots, \; e_m \}$.
%\end{equation}

Similar to works leveraging transformers for mesh triangle generation \citep{chen2024meshanything,tang2024edgerunner}, we directly use the discretized coordinates of our graph vertices and the class indices of attributes as the token indices. A GPT-style transformer $\mathcal{T}$ is then trained autoregressively  to predict tokens sequentially with a Cross Entropy loss, conditioned on image features $z$ extracted from the input RGB image $I$ by a pretrained image encoder \citep{ranzinger2024radio} $ z = \phi(I)$.

\subsection{Autoregressive Generation with MCTS-Guided Sampling}

During inference, we encode an input image $I$ into its features $ z = \phi(I)$, and feed $z$ into the transformer $\mathcal{T}$, which autoregressively generates an output token sequence representing a procedural graph $\gG$. 
While conditioning solely on $z$ allows the model to capture the overall structure of the object,  relying soley on the learned prior can lose accuracy in aligning to finer-grained image details.
To address this, we introduce a Monte Carlo Tree Search (MCTS) guided sampling into our generation process to refine the synthesized token sequence to better align with the image condition.
For alignment guidance, we extract from $I$ its mask $M_I$ (e.g., with SAM~\citep{kirillov2023segment,ren2024grounded}). We then compare $M_I$ with our rendered silhouette $M_{\hat{\gG}}$ of the currently generated $\hat{\gG}$.

% \ANGIE{we should mention here that  we extract a mask (and what machine generated masks we use) for the image alignment guidance so the mask part doesn't come out of nowhere later on}

% \ANGIE{i think i made only minor changes but would be good if you can double check i didn't accidentally make any errors here}

MCTS enables effectively analyzing the most promising choices for the next potential edge in $\gG$ by expanding a search tree through random sampling of the search space, based on our learned prior from $\mathcal{T}$. More specifically, we treat a single edge $e$ comprising $p$ tokens as the minimal search unit and apply MCTS to explore edge candidates that best align with the image condition $I$. 

Our MCTS procedure adopts the standard four-stage procedure: \textbf{selection}, \textbf{expansion}, \textbf{simulation}, and \textbf{reward propagation}\footnote{Referred to as backpropagation in standard MCTS; we use the term reward propagation here to distinguish it from gradient backpropagation.}, which are repeated iteratively.
% \ANGIE{might help to clarify here the terminology for the MCTS search (node, leaf node) vs the procedural graph terminology (edges, can be confusing since the node/edge confusion), can you give the mcts nodes variables with mcts subscripts, and also refer to the edges with their variables? }

Formally, we define states and successor relation of MCTS as follows:
\vspace{0.1cm}
\\
\hspace*{2em} $\circ$ States $S$: each state $s_i \doteq \{ e_1, \; \langle \& \rangle, \; e_2, \; \langle \& \rangle, \; \dots, \; e_i \} \in S$ is a partial graph.
\vspace{0.1cm}
\\
\hspace*{2em} $\circ$ The child state relation is defined by the successor function: $s' \in succ(s_i)$, if there exists \\
\hspace*{2.5em} some $e_{i+1}$ s.t. $s' \doteq \{ e_1, \; \langle \& \rangle, \; e_2, \; \langle \& \rangle, \; \dots, \; e_i, \langle \& \rangle,  e_{i+1} \}$.

% \begin{figure}[t]
% \begin{center}
% \includegraphics[width=0.8\textwidth]{figs/mcts.png}
% \end{center}
% \caption{Overview of our MCTS-guided search to reconstruct a procedural graph that well-aligns to the input image condition. }
% \end{figure}

\begin{figure}[t]
\begin{center}
\includegraphics[width=0.9\textwidth]{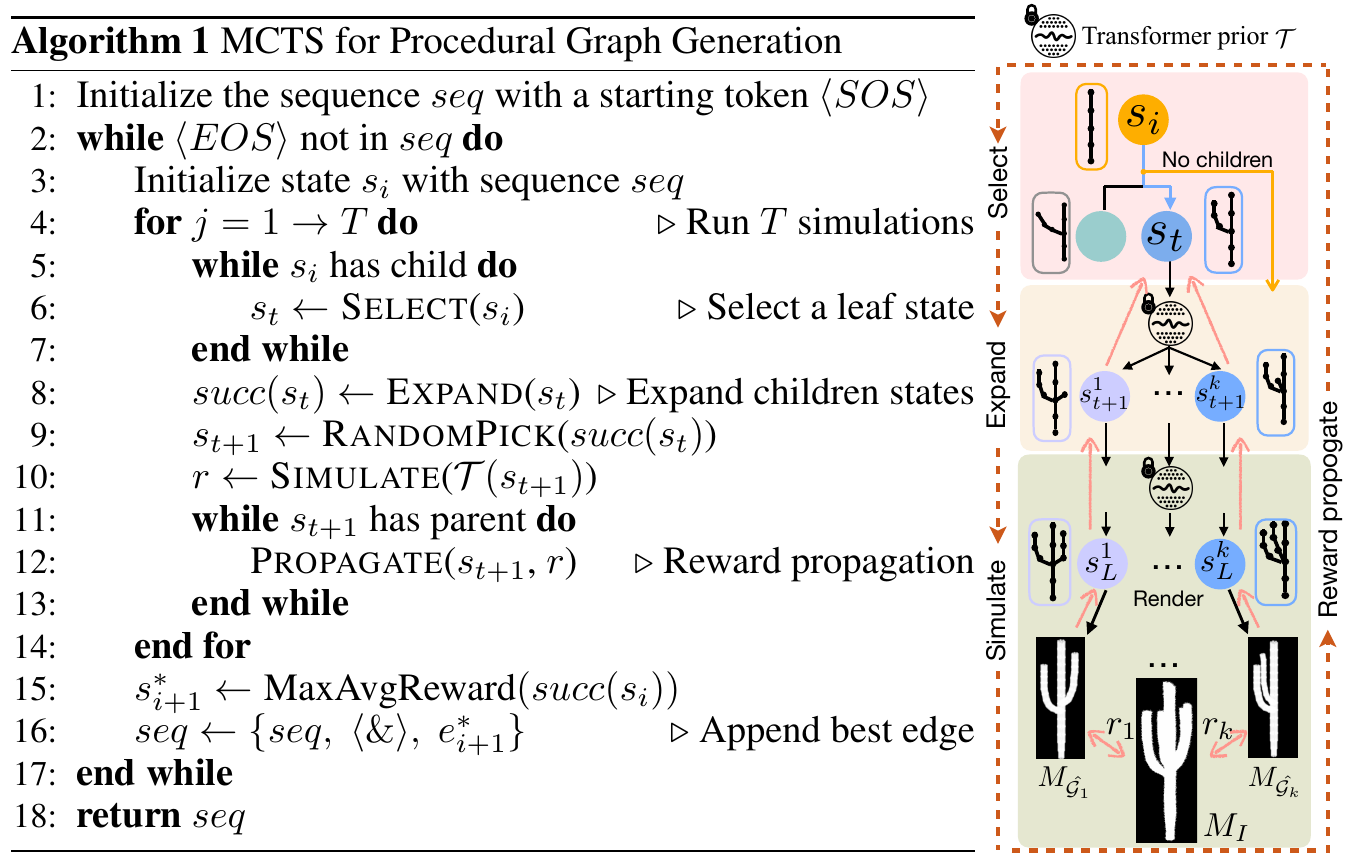}
\end{center}
\vspace{-0.5cm}
\caption{Overview of our MCTS-guided search to reconstruct a procedural graph that well-aligns to the input image condition. }
\end{figure}

%\ANGIE{it would be helpful to connect this more to the terminology that you had for the procedural graph (especially in the simulation part where there are no variable references) and explain more what happens also at the start where there aren't any previously generated edges}

\paragraph{Selection.}
% Starting from the beginning token, the algorithm recursively selects edges along the search tree according to a balance of exploration and exploitation until a ending token is reached. 

Starting from a sequence of previously generated partial graph $seq = s_i$, our goal is to identify a promising leaf state $s^*_{t}$
% $s_{t}\in succ(..succ(s_i))$ 
for simulating future performance. 
Note that at the beginning of the search, $s_{i}$ has no children, and so will be directly selected as the target leaf for expansion. As iterations proceed, however, the number of child states grows exponentially.

While a purely greedy search over all possible leaf states could in principle yield good results, the vast search space makes it computationally infeasible. 
On the other hand, focusing only on states with highest current scores risks getting trapped in local minima due to limited exploration.

To balance exploration and exploitation, we recursively select promising child states $\{s^*_{i+1},s^*_{i+2}, .. \}$ according to an Upper Confidence Bound (UCB) criterion until a leaf state  $s^*_{t}$
is reached,
\begin{equation}
    s_{i+1}^* = \arg\max_{s_{i+1}} \Big( Q(s_i,s_{i+1}) + c \sqrt{\frac{\log N(s_i)}{1 + N(s_i,s_{i+1})}} \Big),
\end{equation}
where $Q(s_{i},s_{i+1})$ is the average reward of choosing candidate child state $s_{i+1} \in succ(s_{i})$, $N(s_{i})$ and $N(s_{i},s_{i+1})$  are the visit counts
of state $s_{i}$ and state $s_{i+1}$, and $c$ is an exploration constant. 
% \ANGIE{state s should be explained or introduced previously}

\paragraph{Expansion.}
At a leaf state $s_t$, new candidate child states $succ(s_t)$  are added to the search tree by predicting the next possible edges.
Given the predicted logits $\ell$ from the transformer prior, we sequentially sample $p$ tokens to form an edge candidate and collect a set of candidates $\mathcal{E} = \{e^1_{t+1},..,e^k_{t+1}\}$ as potential continuations of the sequence, forming a set of new child states 
$succ(s_t) = \{ e_{1:t}, \; \langle \& \rangle, \; e_{t+1} \mid e_{t+1} \in \mathcal{E} \}$.

\paragraph{Simulation \& Reward Propagation.}
After expanding new edges, we simulate their subsequent performance by using the transformer $\mathcal{T}$ to autoregressively generate future tokens up to a predefined length $L$ (e.g., the next 10 edges): $\mathcal{T}(e_{t+1:L} \mid \mathbf{z}) = \prod_{l=t+1}^{L} \mathcal{T}(e_l \mid e_{<l}, \mathbf{z}).$

For each simulation, the quality of the generated sequence is evaluated against the image condition $I$. 
Specifically, edges are represented as cylinders and rendered in real time to obtain a mask $M_{\hat{\gG}}$ of the generated graph $\hat{\gG} = \{e_{1:L} \}$. The overlap between $M_{\hat{\gG}}$ and the mask $M_{I}$ extracted from the input $I$ is then computed to yield a reward score $r$:
% \begin{equation}
%     r = \text{Overlap}(M_{\hat{\gG}}, M_{I}).
% \end{equation}
\begin{equation}
r = \lambda \cdot 
\frac{|M_I \cap M_{\hat{\gG}}|}{|M_{\hat{\gG}}|} 
+ (1 - \lambda) \cdot 
\frac{|M_I \cap M_{\hat{\gG}}|}{|M_I|},
\end{equation}
where $\lambda \in [0,1]$ balances the two overlap ratios.
% \ANGIE{is overlap raw overlap? (vs partial overlap/iou/some other metric)?}
$r$ is then propagated back through all parent states $\{s_{i+1},..., s_t \}$ to update their average reward.

After completing all simulations, the child state of the root state with the highest average reward $s^*_{i+1}$ is selected as the best candidate, and 
its corresponding edge sequence $\{e_{1:i+1} \}$ becomes the new root state for the next iteration. By iteratively applying these four steps, MCTS effectively guides sampling to edge sequences representing procedural graphs more consistent with the image.

\section{Experiments}

In our experimental results, we report quantitative and qualitative evaluations on synthetic procedurally generated data (as described below), where ground truth data is available for evaluation. 
We further demonstrate the applicability of \OURS{} to real images, where no ground truth exists but qualitative results highlight our generalization ability.

\paragraph{Dataset.}
We construct a dataset using procedural generators for both natural and man-made structures. For natural categories, we use Infinigen~\citep{raistrick2023infinite} to generate diverse instances of cacti and trees. 
For trees, this encompasses both standard and pine trees, and we consider two evaluation settings: (i) branches only, which allows accurate assessment of structural alignment, and (ii) with foliage (leafy), which introduces significant occlusion and in a more challenging reconstruction scenario. 
For man-made structures, we adopt a Combinatorial Equilibrium Modeling (CEM) \citep{ohlbrock2020computer} procedural generator to produce bridges. %, which captures regular and engineered design patterns. 

Each generated object is represented by its procedural graph and paired rendered RGB images. %, forming a dataset used for training and evaluation. 
For each category, we generate 10,000 samples (9500/100/400 train/val/test).
%and divide them into training (9500), validation (100) and test (400) subsets. 
We collect different sets of attributes depending on the data category:
\\
%\begin{itemize}
%\item {
\hspace*{2em} $\circ$ Cactus: node attributes include $(x, y, z)$ coordinates and branch radius.
%}
\\
%\item {
\hspace*{2em} $\circ$ Tree: node attributes include $(x, y, z)$ coordinates.
%}
\\
%\item {
\hspace*{2em} $\circ$ Bridge: node attributes include $(x, y, z)$ coordinates and semantic type, while edge attributes
\\ \hspace*{2.5em} include the sign of force, semantic type, and CEM type.
%}
%\end{itemize}

Note that these procedural graphs can then be fed back into the corresponding original procedural generators to produce high-fidelity 3D assets.

% \begin{figure}[t]
% \begin{center}
% \includegraphics[width=0.9\textwidth]{figs/dataset.png}
% \end{center}
% \caption{Examples from our dataset.}
% \label{tab:dataset}
% \end{figure}

\vspace{-0.2cm}
\paragraph{Implementation Details.}
We adopt OPT-350M \citep{zhang2022opt} as our transformer architecture, and the Radio \citep{ranzinger2024radio} pretrained image encoder. For token prediction, we discretize continuous attributes (e.g., coordinates, radii) into 128 classes. For $n$-class categorical attributes, we assign token indices in $[128, 128+n)$ to  avoid overlap with continuous attributes.
During the transformer embedding stage, discretized coordinates and radii are embedded into 1024-dimensional feature vectors. For tokens representing semantic attributes, we use a pretrained CLIP encoder \citep{radford2021learning} to embed their text labels, followed by a linear projection into a 1024-dimensional space. All token embeddings are then concatenated in their original order to form the input sequence to the transformer.
Our model is trained with one A6000 GPU for one day.
%We implemented our approach using PyTorch \citep{paszke2019pytorch}.

\subsection{Image to 3D reconstruction}

% basline: TRELLIS Wonder3D TreeDFusion(only faliaged tree)
\paragraph{Baselines.}
We compare our image to 3D reconstruction results with state-of-the-art  image to 3D generative methods TRELLIS \citep{xiang2025structured} and Wonder3D \citep{long2024wonder3d}. We further compare with the domain-specific method TreeDFusion \citep{lee2024tree} on the leafy tree category. %, a baseline trained specifically on tree datasets. 

% table one (3d + 2d) done
\vspace{-0.2cm}
\paragraph{Evaluation metrics.}
Following prior work \citep{lee2024tree}, we evaluate both 3D and 2D metrics: 1) Chamfer Distance between point clouds sampled from ground-truth meshes and reconstructed meshes (direct outputs from baselines; reconstructed from generated graphs via the corresponding procedural generator for our method), 2) LPIPS and 3) CLIP similarity between the input RGB images and renderings of the reconstructions.

\begin{figure}[t]
\vspace{-0.4cm}
\begin{center}
\includegraphics[width=0.98\textwidth]{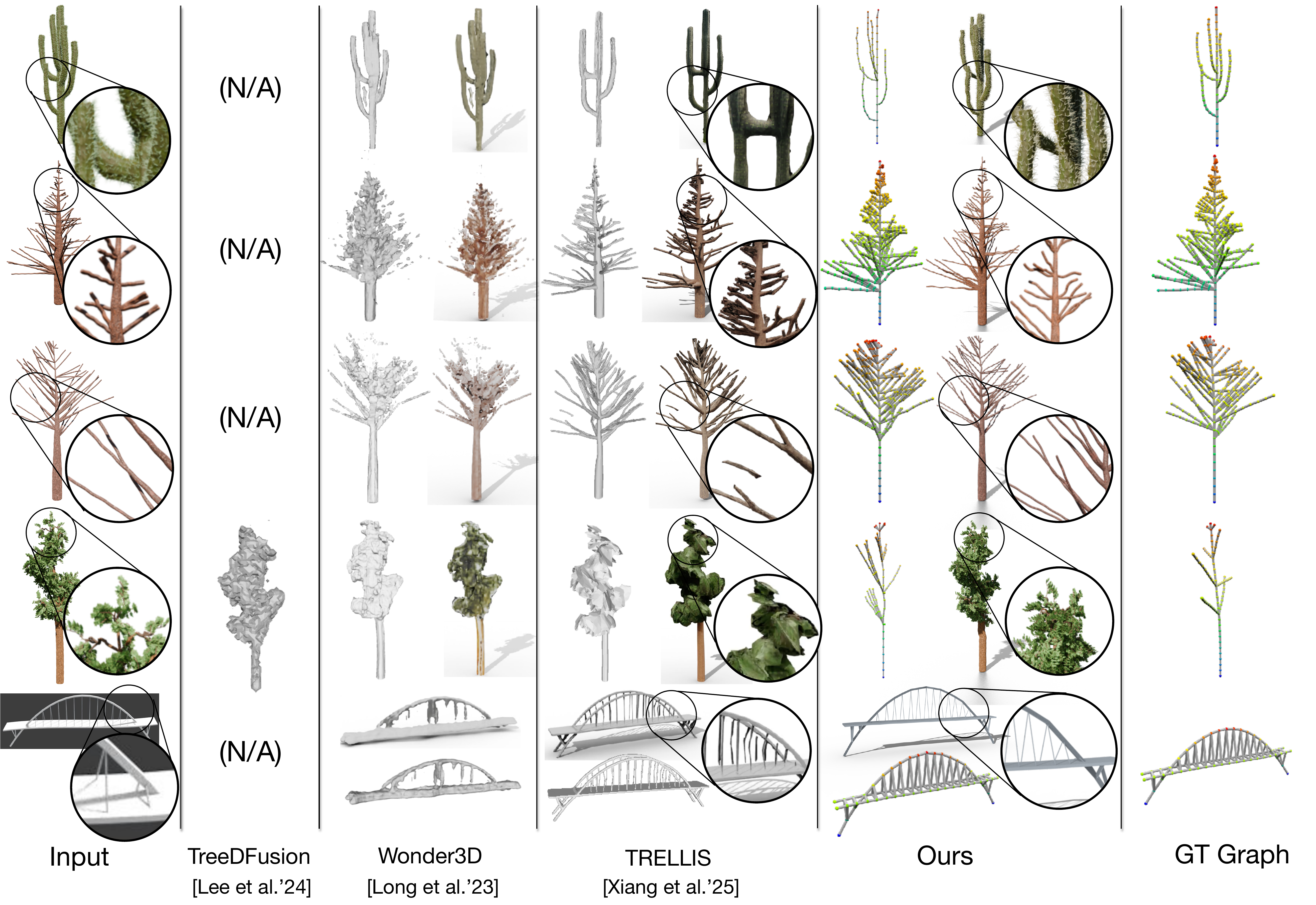}
\end{center}
\vspace{-0.2cm}
\caption{Qualitative comparison with state-of-the-art 3D generative models TRELLIS and Wonder3D. \OURS{} achieves higher-fidelity reconstructions on all categories.
%\ANGIE{need zoomins here, since the trellis tree and bridge look pretty good at this scale}
}
\label{fig:comparison}
\end{figure}

As shown in Tab.~\ref{tab:comparison}, our method outperforms all baselines across all categories, demonstrating superior geometric accuracy, perceptual fidelity, and semantic alignment with the input images. Qualitative results in Fig.~\ref{fig:comparison} further show that our reconstructions better align with input images. 
In particular, Wonder3D often fails to reconstruct fine-grained structures (e.g., dense or thin elements), while TRELLIS can capture the overall structure but struggles with local geometric fidelity, leading to distorted or physically implausible details such as broken branches and bridge cables. In contrast, our procedural graph representation inherently encourages structural correctness during reconstruction. This allows \OURS{} to preserve both global structure and fine-grained details, producing reconstructions that are  much more consistent in both structure and geometry.

\begin{table}[t]
\vspace{-0.4cm}
\caption{Comparison with TRELLIS \citep{xiang2025structured}, Wonder3D \citep{long2024wonder3d} and TreeDFusion \citep{lee2024tree}. Our approach achieves superior performance across all categories, in both 3D and 2D metrics.}
\begin{center}
% \resizebox{.8\linewidth}{!}{
\begin{tabular}{c c c c c}
\hline
Category & Method & CD $\downarrow$ & LPIPS $\downarrow$ & CLIP-Sim $\uparrow$ \\
\hline
\multirow{3}{*}{Cactus} 
 & TRELLIS   & 0.0456 & 0.195 & 0.9222 \\
 & Wonder3D  & 0.0375 & 0.141 & 0.9029 \\
 & Ours      & \textbf{0.0297} & \textbf{0.097} & \textbf{0.9268} \\
\hline
\multirow{3}{*}{Tree} 
 & TRELLIS   & 0.0459 & 0.158 & 0.9644 \\
 & Wonder3D  & 0.0479 & 0.212 & 0.9013 \\
 & Ours      & \textbf{0.0265} & \textbf{0.081} & \textbf{0.9769} \\
\hline
\multirow{3}{*}{Leafy Tree} 
 & TRELLIS   & 0.0775 & 0.199  & 0.9348 \\
 & Wonder3D  & 0.0745 & 0.214 &  0.8748 \\
 & TreeDFusion  & 0.0991 & 0.356 & 0.8543 \\
 & Ours      & \textbf{0.0648} & \textbf{0.168} & \textbf{0.9493} \\
\hline
\multirow{3}{*}{Pine tree} 
 & TRELLIS   & 0.0508 & 0.204 & 0.9496 \\
 & Wonder3D  & 0.0599 & 0.254 & 0.8834 \\
 & Ours      & \textbf{0.0302} & \textbf{0.079} & \textbf{0.9680} \\
\hline
\multirow{3}{*}{Bridge} 
 & TRELLIS   & 0.0363 & 0.156 & 0.9175 \\
 & Wonder3D  &  0.1097 & 0.302 & 0.7763 \\
 & Ours      & \textbf{0.0141} & \textbf{0.052} & \textbf{0.9820} \\
\hline
\end{tabular}
% }
\end{center}
\label{tab:comparison}
\end{table}

% figure one (comparison, one example per class)

% table two: foliaged tree (ours+3 baselines, matrix: 2d)
% figure two (comparison, 1 or 2 examples)

% \begin{figure}[t]
% \begin{center}
% \includegraphics[width=0.95\textwidth]{figs/tree_full.png}
% \end{center}
% \caption{Qualitative comparison with baselines on leafy trees. In this challenging scenario, our procedural graph basis and MCTS refinement enable more faithful reconstructions of the input image.}
% \label{fig:comparison_tree}
% \end{figure}

\subsection{Ablations}

To evaluate the effectiveness of our representation in capturing structural correctness, we conduct ablation studies focusing on the impact of different input modalities (mask vs. RGB), tokenization ordering (DFS vs. BFS), and our use of MCTS guided sampling.

\vspace{-0.2cm}
\paragraph{Effect of input image modality and tokenization order}
We first study the influence of different input modalities and tokenization strategies on reconstruction quality. 
% For clarity, we evaluate performance directly on graphs, measuring the graph Chamfer Distance CD$_{\gG}$ and topological similarity between ground-truth and predicted graphs.
% \ANGIE{the justification for different evaluation measures here kind of begs the question what the results on the standard CD / LPIPS / CLIP metrics would be, do we have those as well?}
We consider only using extracted image masks as input rather than RGB along with extracted masks. 
As shown in Tab.~\ref{tab:aba1}, the RGB inputs tend to provide more information (particularly regarding occlusions), leading to better performance.
We also compare depth-first search (DFS) and breadth-first search (BFS) orders for graph tokenization. While both orders achieve similar levels of topological consistency, DFS exhibits a clear advantage in preserving spatial fidelity.

\begin{table}[t]
\vspace{-0.2cm}
\caption{Ablation study of our design choices on the tree category. The results show that both incorporating RGB information and adopting an appropriate tokenization order significantly improve structural consistency and geometric fidelity in reconstruction.}
\centering
\begin{tabular}{c | c c c c c }
\hline
 Categroy &      Setting      & CD $\downarrow$ & Topo-Sim $\uparrow$ & LPIPS $\downarrow$ & CLIP-Sim $\uparrow$\\
\hline
\multirow{4}{*}{Tree} 
 & mask\_input & 0.0485 & 0.958 & 0.166 &  0.9319\\
 & rgb\_input  & \textbf{0.0265} & \textbf{0.971} & \textbf{0.081} & \textbf{0.9769} \\
 \cline{2-6} 
 & \rule{0pt}{2.0ex}BFS\_order    & 0.0336 & \textbf{0.972} & 0.137 & 0.9701\\
 & DFS\_order    & \textbf{0.0265} & 0.971 & \textbf{0.081} & \textbf{0.9769} \\
\hline
\end{tabular}
\label{tab:aba1}
\end{table}

\vspace{-0.2cm}
\paragraph{Effect of MCTS guided sampling}
To assess the effectiveness of our MCTS-guided sampling in capturing accurate fine-grained local detail, we compare to a baseline variant of our approach  conditioned solely on RGB images without MCTS. 
%Following the same evaluation settings used for baseline comparisons, we report performance across mesh quality, graph structure and rendering fidelity.
As shown in Tab.~\ref{tab:cond_mcts}, incorporating MCTS consistently improves reconstruction quality across all categories.
These improvements are particularly pronounced in challenging cases with fine-grained details, where direct autoregressive generation often fails to capture local geometry accurately.
However, in the case of leafy trees, the mask becomes heavily expanded due to occlusion, which can cause MCTS to generate additional branches, leading to a slight drop in topological similarity.
For real images, as shown in Sec.~\ref{subsec:realimages}, our MCTS guided sampling improves over conditional generation only on leafy trees, as conditional generation suffers more from a larger domain gap.

\begin{table}[t]
\vspace{-0.5cm}
\caption{Ablation study of our MCTS guided sampling vs. image-conditioned generation only. The results demonstrate that incorporating MCTS significantly improves consistency with the inputs.
}
\begin{center}
\resizebox{.8\linewidth}{!}{
\begin{tabular}{c c c c c c}
\hline
Category & Method & CD $\downarrow$ & Topo-Sim $\uparrow$ & LPIPS $\downarrow$ & CLIP-Sim $\uparrow$ \\
\hline
\multirow{2}{*}{Cactus}  
& w/o MCTS   & 0.0311 & 0.985  & 0.108 & 0.9268 \\
& w MCTS (Ours)    & \textbf{0.0297} & \textbf{0.987}  & \textbf{0.097} & \textbf{0.9268} \\
\hline
\multirow{2}{*}{Tree}  
& w/o MCTS   & 0.0312 & 0.970 & 0.116 & 0.9533 \\
& w MCTS (Ours)         & \textbf{0.0265} & \textbf{0.971}  & \textbf{0.081} & \textbf{0.9769} \\
\hline
\multirow{2}{*}{Leafy Tree} 
& w/o MCTS  & 0.0653 & \textbf{0.962} & \textbf{0.152} & 0.9424 \\
& w MCTS (Ours) &  \textbf{0.0648}  &  0.948  &    0.168   &   \textbf{0.9493}    \\
\hline
\multirow{2}{*}{Pine tree} 
& w/o MCTS & 0.0313 & 0.980 & 0.091 & 0.9673 \\
& w MCTS (Ours)       & \textbf{0.0302} & \textbf{0.983}  & \textbf{0.079} & \textbf{0.9677} \\
\hline
\multirow{2}{*}{Bridge}
 & w/o MCTS   & 0.0164 & 0.996 & 0.058 & 0.9785 \\
 & w MCTS (Ours)         & \textbf{0.0141} & \textbf{0.998}  & \textbf{0.052} & \textbf{0.9820} \\
\hline
\end{tabular}
}
\end{center}
\vspace{-0.4cm}
\label{tab:cond_mcts}
\end{table}

\vspace{-0.2cm}
\subsection{Generalization to Real-World Images}
\label{subsec:realimages}
\vspace{-0.2cm}

\begin{figure}[b]
\vspace{-0.4cm}
\begin{center}
\includegraphics[width=0.9\textwidth]{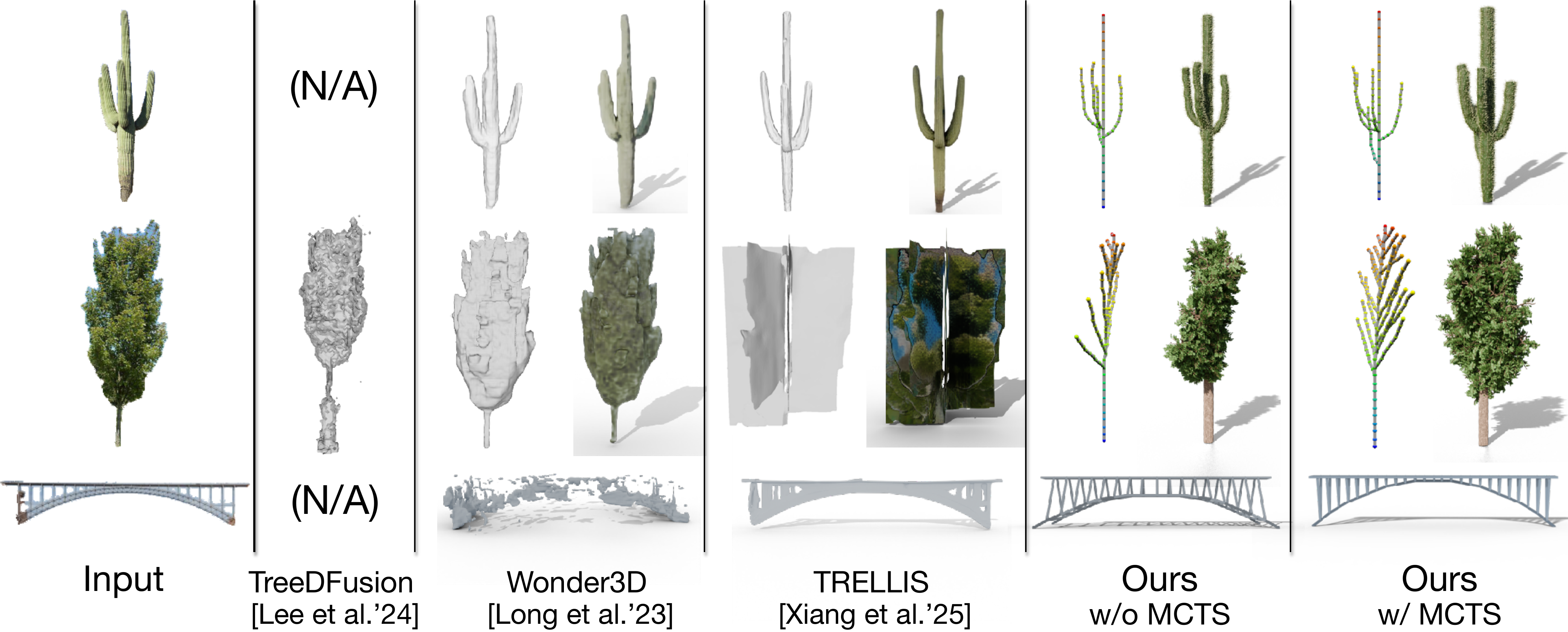}
\end{center}
\vspace{-0.5cm}
\caption{Comparison with baselines on real-world cactus, leafy tree, and bridge images. The results demonstrate that our model generalizes effectively to real-world data, despite being trained only on synthetic data.}
\label{fig:real}
\end{figure}

We further test our method's ability to generalize to real-world photos collected from the internet. We employ Grounded-SAM~\citep{ren2024grounded} to extract the mask of the main object, and use the masked image as input. 
As shown in Fig.~\ref{fig:real}, while baselines can capture overall structures of the target objects, they struggle to recover fine-grained details.
In contrast, our method produces more physically plausible and topologically similar structures. Furthermore, incorporating MCTS guidance improves the alignment between the generated graphs and the target image, yielding more consistent reconstructions than our variant without MCTS.
\vspace{-0.4cm}

% \begin{figure}[t]
% \begin{center}
% \includegraphics[width=0.9\textwidth]{figs/real_tree.png}
% \end{center}
% \caption{Comparison with baselines on real-world trees. For a clear evaluation of both visual fidelity and geometric accuracy, we present results as textured meshes as well as pure geometry (untextured meshes).}
% \label{fig:real_tree}
% \end{figure}

% cactus bridge tree

\section{Conclusion}
\vspace{-0.3cm}
We have presented \OURS{}, a new approach for reconstructing 3D objects from RGB images by synthesizing compact procedural graph abstractions, which then can be decoded with a procedural generator into high-fidelity 3D assets. 
Our method trains a transformer to approximate the landscape of the procedural generator in the procedural graph space, using an edge-based tokenization strategy to represent procedural graphs as sequences. 
Unlike existing methods that rely solely on a learned prior for image-to-3D reconstruction, we incorporate an MCTS-guided search over the transformer prior, enabling reconstructions that more faithfully capture local structures of the input image. 
This yields higher-fidelity synthesis, and we hope this highlights the potential of alternative representations in 3D content creation.

\paragraph{Acknowledgments.}
This project was supported by the ERC Starting Grant SpatialSem (101076253) and the TUM Georg Nemetschek Institute Artificial Intelligence for the Built World.

\bibliography{iclr2026_conference}
\bibliographystyle{iclr2026_conference}

% %%%%%%%%% appendix 
\newpage
\appendix
\section{Appendix}

\subsection{Additional Qualitative results}

\begin{figure}[h]
\begin{center}
\includegraphics[width=0.98\textwidth]{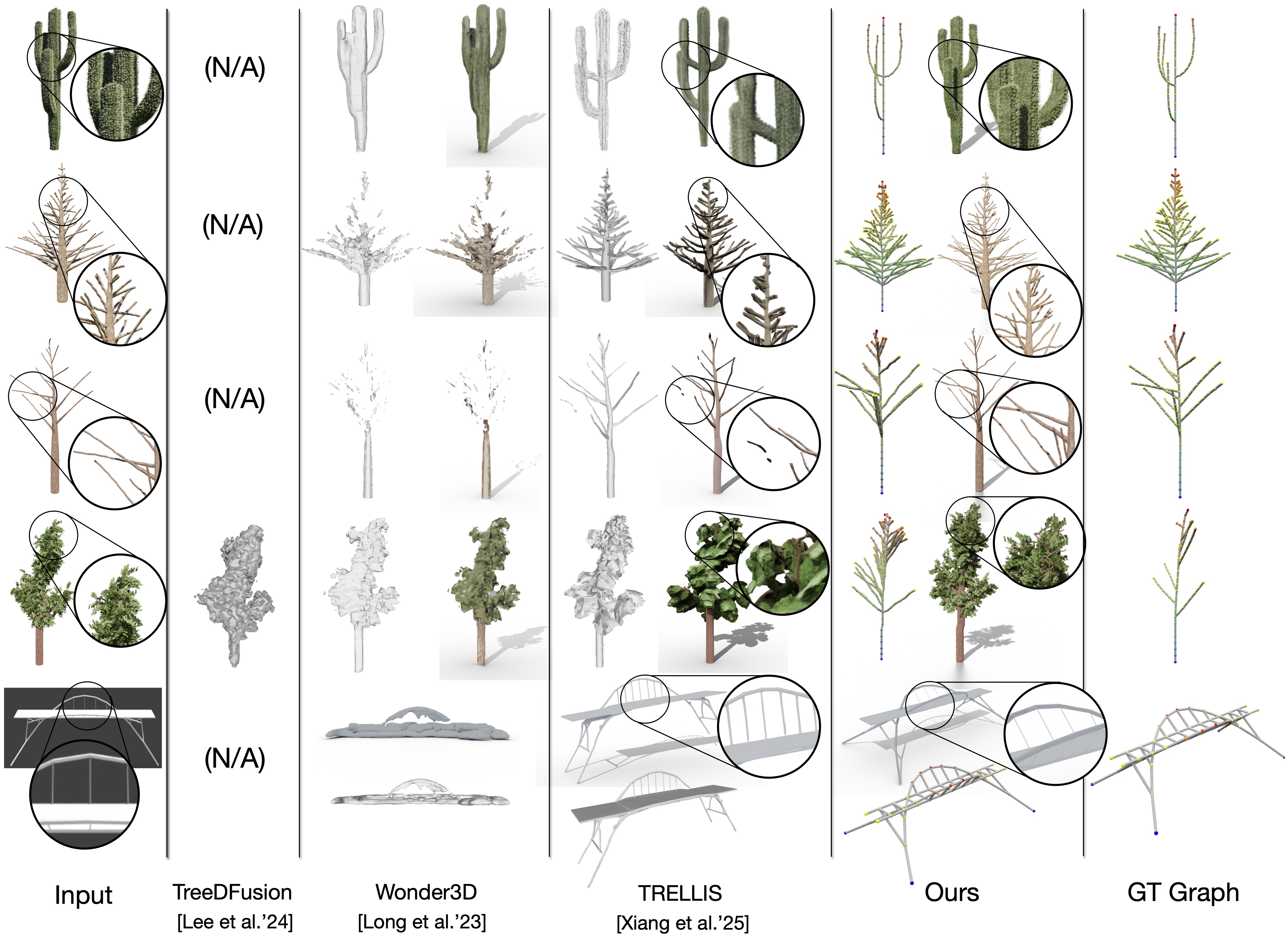}
\end{center}
\vspace{-0.5cm}
\caption{Additional qualitative comparisons with state-of-the-art 3D generative models TRELLIS and Wonder3D. \OURS{} achieves improved reconstruction quality and consistency with the input images. 
%\ANGIE{need zoomins here, since the trellis tree and bridge look pretty good at this scale}
}
\label{fig:sup-comparison}
\end{figure}

\begin{figure}[t]
\begin{center}
\includegraphics[width=0.95\textwidth]{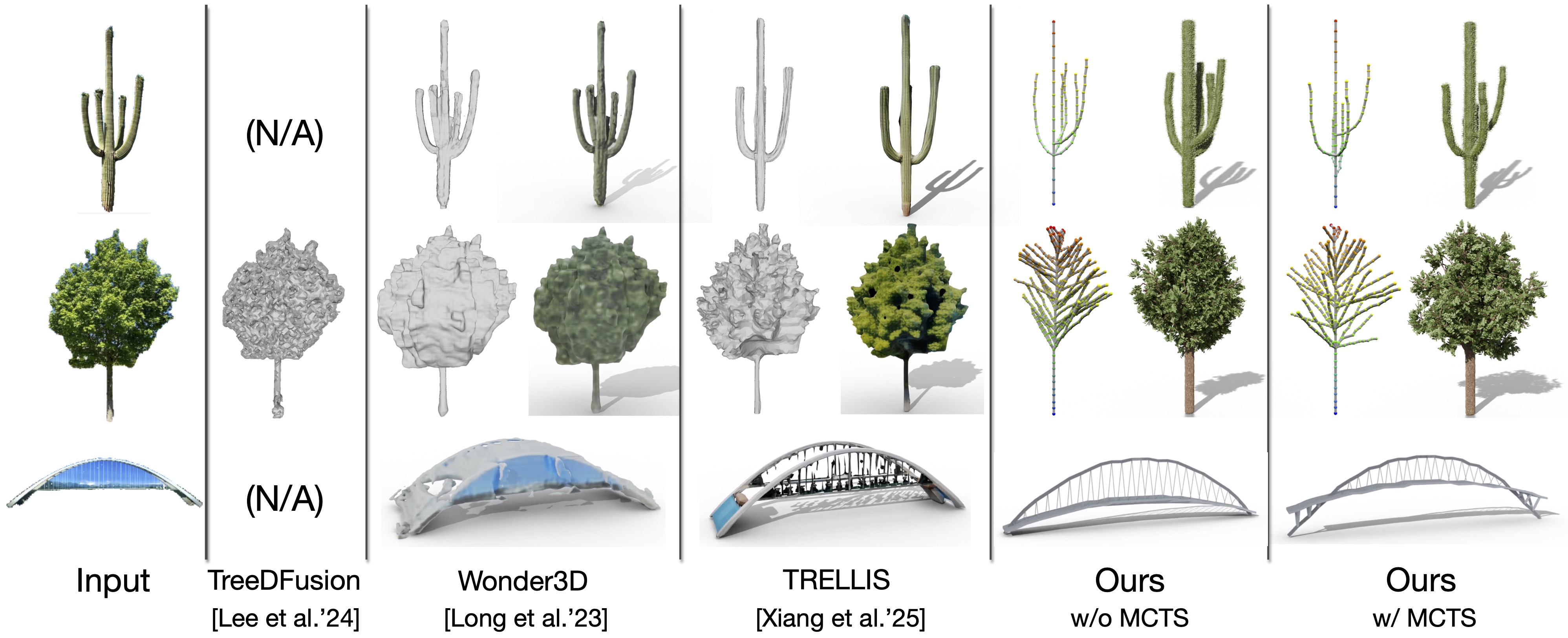}
\end{center}
\caption{Additional qualitative comparison with state-of-the-art 3D generative models TRELLIS and Wonder3D on real world images. Notably, the cactus and bridge inputs differ significantly from the training data, yet our method is still able to reconstruct objects with similar topology.
}
\label{fig:sup-real}
\end{figure}

\begin{figure}[h]
\begin{center}
\includegraphics[width=0.98\textwidth]{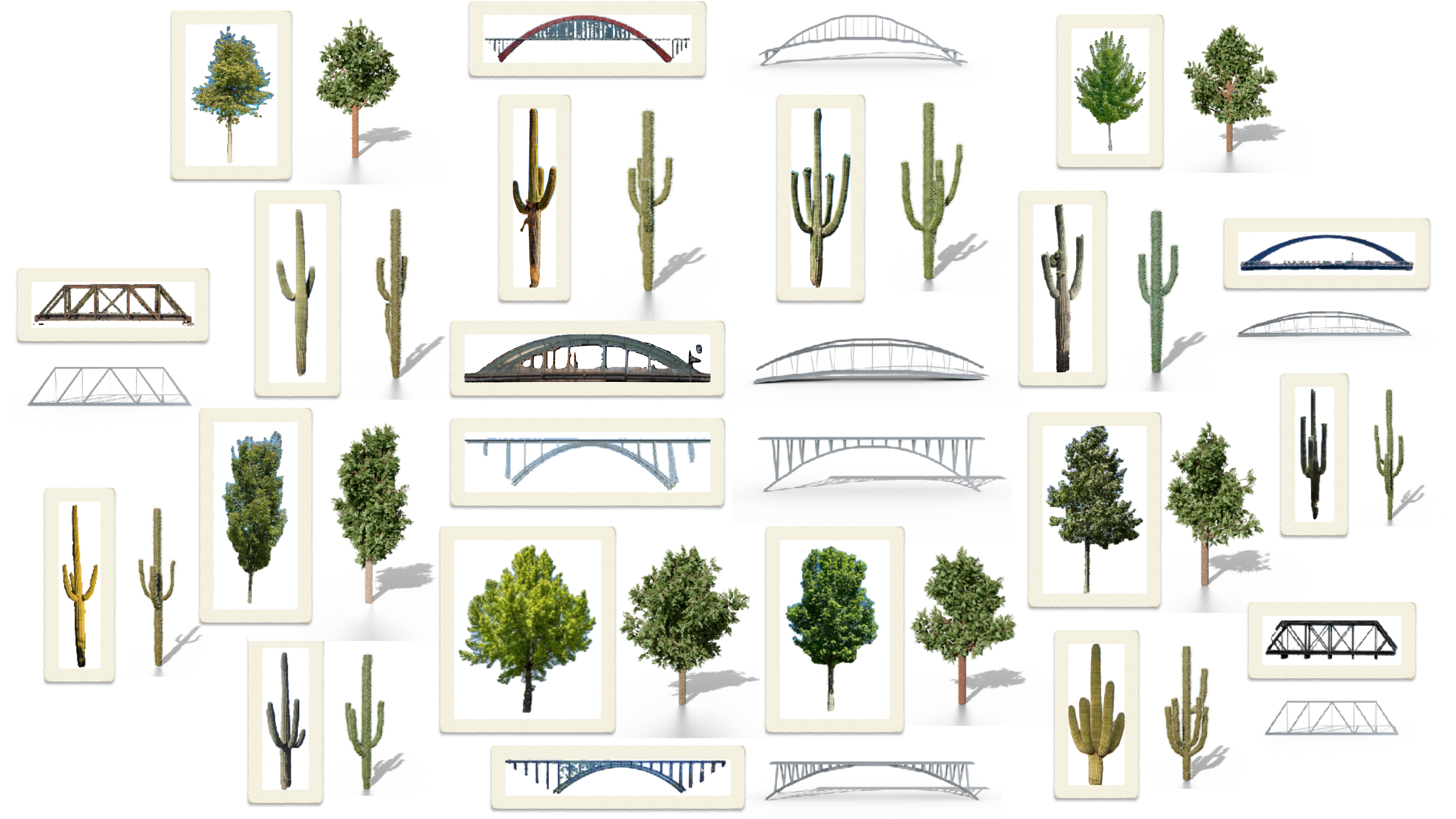}
\end{center}
% \vspace{-0.5cm}
\caption{Gallery of real-world reconstructions with procedural graphs.  Across a wide variety of inputs, our approach consistently generates high-fidelity 3D reconstructions that are both structurally coherent and geometrically detailed.
}
\label{fig:real-gallery}
\end{figure}

To further demonstrate the performance of our model, we provide additional qualitative results in Fig.~\ref{fig:sup-comparison} on synthetic data, Fig.~\ref{fig:sup-real} and Fig.~\ref{fig:real-gallery} on real-world images. The synthetic examples highlight the model’s ability to reconstruct both global structure and fine-grained geometric details across diverse categories. The real-world examples show that our method generalizes beyond the synthetic training domain, producing structurally consistent and geometrically faithful reconstructions despite the presence of noise, occlusion, and domain shift.

\subsection{dataset}
\begin{figure}[h]
\begin{center}
\includegraphics[width=0.9\textwidth]{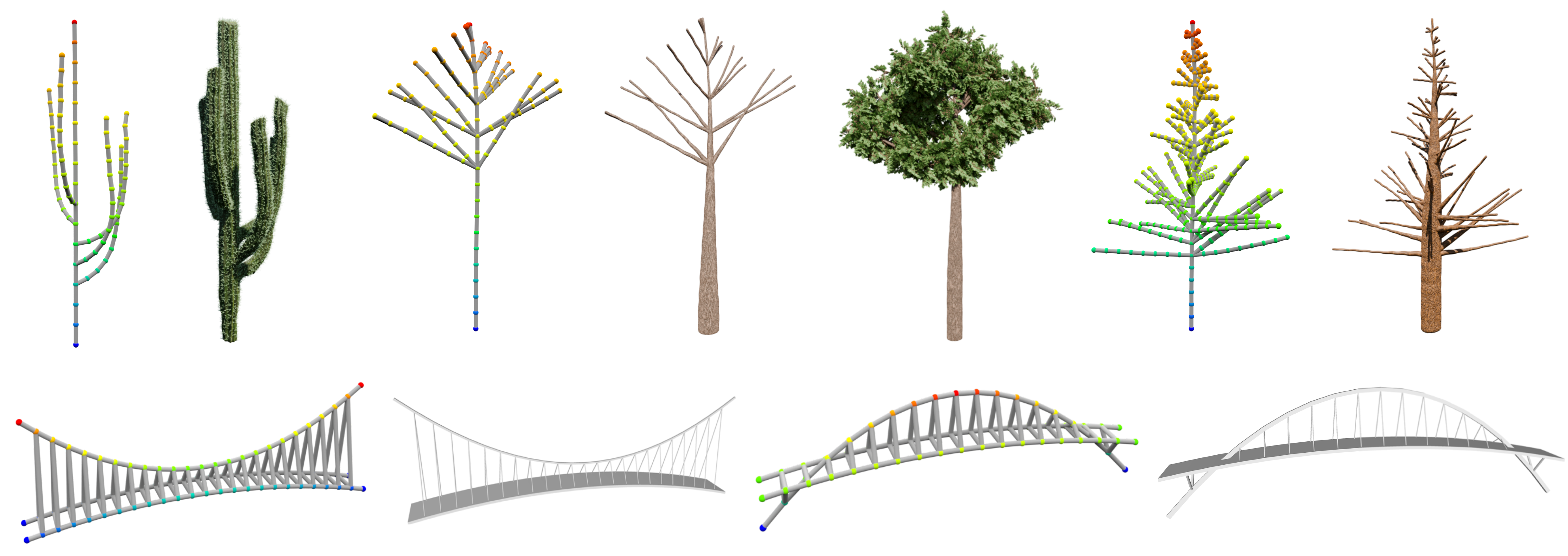}
\end{center}
\caption{Examples of data samples from our dataset, which inclues cacti, trees, pine trees, and bridges. The dataset spans a wide range of structural complexities, from simple graphs with tens of vertices to highly complex graphs with several hundred vertices.
}
\label{fig:dataset}
\end{figure}

We construct our dataset with categories of varying complexity in their procedural graphs. Cacti tend to be simpler, with approximately 30–100 vertices in their graphs. Bridge graphs contain 20–140 vertices, while standard trees range from 50–400 vertices. Pine trees are the most complex, with 300–600 vertices. This variation in structural complexity allows us to systematically evaluate the scalability and robustness of our method across both simple and highly complex objects. Based on these procedural graphs, we use the corresponding procedural generators to synthesize full 3D assets with geometry and realistic textures. As illustrated in Fig.~\ref{fig:dataset}, our dataset covers diverse categories and structural complexities, providing a challenging benchmark for evaluation.

\subsection{Inference Time}
Introducing MCTS at inference time  improves our reconstruction performance and alignment with the input images, though it comes at the cost of increased inference time, particularly for complex categories with multiple plausible solutions (e.g., tree and pine tree, due to strong self-occlusions), as shown in Tab.~\ref{tab:inference}. To help improve performance, we adopt a dynamic expansion strategy that adjusts the number of simulations based on the branching factor. Specifically, at each expansion step of the current state, given the predicted logits from the transformer prior, we apply top-$k$ top-$p$ sampling with $k=50$ and $p=0.95$ to propose $n$ candidate edges for sequence continuation. If $n=1$, the child state is selected directly and simulations are skipped. Otherwise, we perform $n \times \textit{ratio}$ simulations, where the ratio controls the trade-off between computational cost and search quality. 
Our approach remains relatively unoptimized for speed, however, and could be substantially accelerated through parallelizing simulations across multiple GPUs or CPUs.
%This strategy substantially reduces search time compared with the traditional MCTS approach. Looking ahead, inference could be further accelerated through parallelizing simulations across multiple GPUs or CPUs.

\begin{table}[h]
\centering
\caption{Comparison of the average inference time per sample for conditional generation (w/o MCTS) and MCTS across different categories.}
\begin{tabular}{l|cc}
\hline
\textbf{Category} & \textbf{w/o MCTS} & \textbf{w/ MCTS} \\
\hline
Cactus   & 11s & $<$1min \\
Tree     & 24s & 24min \\
Pinetree & 48s & 42min \\
Bridge   & 31s & 2min \\
\hline
\end{tabular}
\label{tab:inference}
\end{table}

\subsection{Limitations}
While our \OURS{} approach demonstrates strong potential as a compact and geometry-faithful representation for 3D content creation, several limitations remain. 
Since our approach builds on procedural generators for data, we focus on categories where such generators are available.
Additionally, while our MCTS-guided sampling improves consistency with input images, it also introduces additional computational cost compared to conditional feedforward generation only.
%relies heavily on the corresponding procedural generators, the overall reconstruction quality is bounded by the fidelity and expressiveness of these generators. Additionally, the generalization ability may be affected when there is a noticeable domain gap between the synthetic training data and real-world inputs. These limitations, however, are not fundamental and could be alleviated by incorporating more diverse training data or integrating domain adaptation techniques in future work.

% \subsection{LLM Usage}
% Grammar and phrasing polish were improved in this paper with the help of LLMs.

\end{document}